\documentclass[conference]{IEEEtran}

\usepackage{cite}
\usepackage{amsmath,amssymb,amsfonts}
\usepackage{algorithm}
\usepackage[noend]{algpseudocode}
\usepackage{graphicx}
\usepackage{textcomp}
\usepackage{xcolor}
\usepackage{url}
\usepackage{enumitem}

\DeclareUnicodeCharacter{2009}{\,} 
\DeclareUnicodeCharacter{202F}{\,} 
\DeclareUnicodeCharacter{2248}{\ensuremath{\approx}} 

\def\BibTeX{{\rm B\kern-.05em{\sc i}\kern-.025emB\kern-.08em
    T\kern-.1667em\lower.7ex\hbox{E}\kern-.125emX}}

\title{Paged Attention Meets FlexAttention: Unlocking Long-Context Efficiency in Deployed Inference}

\author{
\IEEEauthorblockN{Thomas Joshi, Herman Saini, Neil Dhillon}
\IEEEauthorblockA{\textit{Columbia University} \\
Email: \{ttj2108, hss2173, nsd2147\}@columbia.edu}
\IEEEauthorblockN{Antoni Viros i Martin, Kaoutar El Maghraoui}
\IEEEauthorblockA{\textit{IBM}}
}

\begin{document}

\maketitle

\begin{abstract}
Large Language Models (LLMs) encounter severe memory inefficiencies during long-context inference due to conventional handling of key–value (KV) caches. In this work, we introduce a novel integration of PagedAttention with PyTorch's FlexAttention, addressing internal fragmentation and inefficiencies associated with monolithic KV cache allocations. Implemented within IBM's Foundation Model Stack (FMS), our fused attention kernel efficiently gathers scattered KV data. Our benchmarks on an NVIDIA L4 GPU (24GB) demonstrate significantly reduced inference latency, growing only linearly ($\sim$2$\times$) with sequence length from 128 to 2048 tokens when utilizing a global KV cache, compared to exponential latency increases without caching. While peak memory usage remains largely unchanged for single-step evaluations (dominated by model weights and activations), paged attention causes minimal incremental memory usage, observable only at sequence lengths exceeding 2048 tokens due to its power-of-two cache allocations. We open‑source the full implementation and discuss its implications for future long‑context model deployment.
\end{abstract}


\section{Introduction}\label{sec:intro}

Transformer-based large language models (LLMs) have catalyzed dramatic progress in natural-language understanding and generation.  Yet when these models are deployed in real-world inference workloads—chat assistants, document summarizers, or code-completion engines—they encounter a stubborn systems bottleneck: \emph{memory usage grows linearly with the length of the input context and the number of generated tokens}.  Each new token appends key and value (\textit{KV}) vectors for \emph{every} attention layer, creating a rapidly expanding KV--cache.  For state-of-the-art models with thousands–to–hundreds-of-thousands-token contexts, the KV cache alone can occupy tens of gigabytes of GPU memory~\cite{dao2022flashattention,kwon2023pagedattention}.  Once device memory is exhausted, inference engines must either reject new requests or page data to host memory, incurring orders-of-magnitude slowdowns and violating user latency budgets.

\vspace{0.5em}
\noindent\textbf{Internal \& external fragmentation.}
To avoid expensive GPU reallocations, most production systems \emph{pre-allocate} a contiguous KV buffer for each request sized to the maximum supported sequence length.  If the \emph{actual} sequence is shorter, the unused tail of that buffer is "dead" memory.  Empirical studies report 60–80\% average waste for mixed-length batches in popular inference servers~\cite{kwon2023pagedattention,deepspeed2023blog}.  When many requests are multiplexed on one accelerator, the fixed buffers also introduce \emph{external} fragmentation: once the GPU address space is carved into these immutable blocks, the gaps between them are rarely the right shape for the next request, even if the total free space would suffice. 

\begin{figure}[t]
  \centering
  \includegraphics[width=0.8\linewidth]{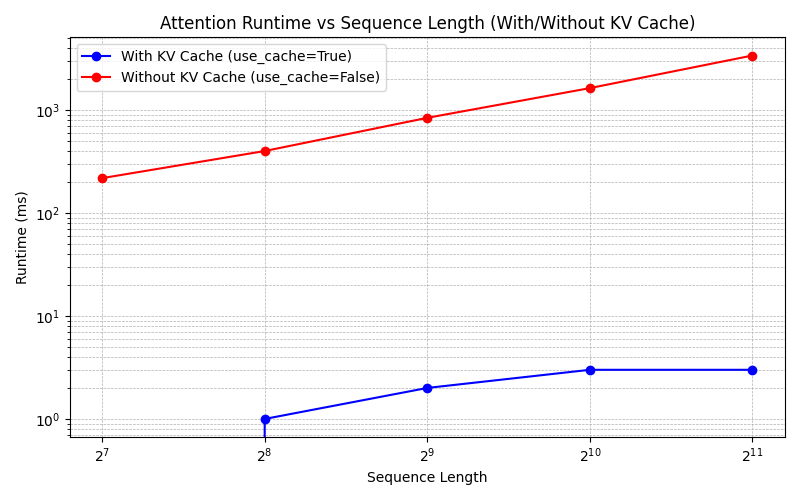}
  \caption{\textbf{Inference latency versus sequence length with PagedAttention on an NVIDIA L4 GPU (24 GB).} Latency grows roughly linearly ($\sim$2$\times$ across 128–2048 tokens) when the global KV cache is enabled, while disabling the cache leads to an exponential increase ($\sim$10$\times$ per doubling). The cached KV tensors eliminate redundant computation, sustaining high throughput for autoregressive generation workloads.}
  \label{fig:runtime_paged_l4}
\end{figure}

\vspace{0.5em}
\noindent\textbf{Practical impact.}
Memory waste throttles \emph{throughput}: the accelerator runs out of capacity long before compute saturates, forcing requests to queue.  It also inflates \emph{latency}: when the cache spills to CPU, token generation rate can plummet by $10{\times}$ or more.\,(see Figure~\ref{fig:runtime_paged_l4}).  These effects limit product features, for example, long-form document reasoning, retrieval-augmented generation (RAG) with large evidence windows, or multiuser batching for cost efficiency.  Industry has begun to respond: the \texttt{vLLM} server demonstrates that \emph{PagedAttention} can sustain 32k-token contexts at high speed~\cite{kwon2023pagedattention}; Google reports "infinite context" prototypes for PaLM; Microsoft's \texttt{vAttention} exploits GPU demand paging to similar ends~\cite{prabhu2024vattention}.  All point to the same conclusion: eliminating KV fragmentation is pivotal for the next generation of LLM services.

\subsection{Problem Statement}
We ask: \textit{How can we serve long-context LLMs on commodity GPUs \underline{without} prohibitive memory overhead or invasive model changes?}  

Two intertwined challenges arise:

\begin{enumerate}
    \item \textbf{Dynamic, fine-grained memory management.}  The KV cache must grow on-demand for many independent sequences, reclaim space instantly when a sequence finishes, and share identical prefixes across requests—all with \emph{constant-time} allocation to stay off the critical path.
    \item \textbf{Compute-efficient attention over non-contiguous memory.}  Once KV tensors are scattered in memory pages, the attention kernel must gather them \emph{without} extra copies and with throughput comparable to highly optimized fused kernels such as FlashAttention \cite{dao2022flashattention}.
\end{enumerate}

\subsection{Objectives and Scope}
This paper targets inference (decode-time) optimization for decoder-only LLMs executed on single- or multi-GPU servers.

Our goals are:

\begin{itemize}[leftmargin=*]
    \item \textbf{Zero-waste KV cache.}  Achieve $<\!5\%$ memory overhead relative to the theoretical minimum, independent of batch composition.
    \item \textbf{Open, reproducible implementation.}  Provide source code, unit tests, and benchmarking scripts in a public GitHub repository for the community.
    \item \textbf{Sustained throughput.}  Deliver equal or higher tokens-per-second than standard fused attention for contexts up to \textbf{32 k} tokens on commodity GPUs (T4/L4).
    \item \textbf{Drop‑in deployability.}  Integrate into IBM's Foundation Model Stack (FMS) via configuration flags, requiring \emph{no} model re‑training or architecture edits.
\end{itemize}

\vspace{0.5em}
\noindent\textbf{Approach overview.}
We adopt \emph{PagedAttention}—an OS-inspired scheme that partitions the KV cache into fixed-size pages and tracks them with a block table~\cite{kwon2023pagedattention}.  To execute attention over these non-contiguous pages at near-FlashAttention speed, we leverage PyTorch~2.\,x's recently released \texttt{FlexAttention} API, which JIT-fuses user-provided masking and indexing logic into a single CUDA kernel.  Our contributions are threefold:

\begin{enumerate}
    \item A \emph{KV-page manager} that performs lock-free allocation, deallocation, and prefix sharing in $O(1)$ time.
    \item A \emph{FlexAttention mask} that maps each query to exactly its sequence-local pages, enabling coalesced memory reads and preserving numerical equivalence with standard attention.
    \item A \emph{system-level integration} into FMS's multi-head attention module, validated on LLaMA-7B across T4 and L4 GPUs.
\end{enumerate}

Extensive experiments demonstrate significant reductions in inference latency, scaling linearly rather than exponentially with sequence length, and minimal changes in peak memory usage during single-step evaluations, all while maintaining identical perplexity.

The remainder of the paper is organized as follows.  Section~\ref{sec:related} reviews related work.  Section~\ref{sec:method} details the paged KV manager and FlexAttention kernel.  Section~\ref{sec:results} presents empirical results.  Section~\ref{sec:discussion} discusses limitations and future directions, and Section~\ref{sec:conclusion} concludes.

\section{Related Work}\label{sec:related}

\subsection{Review of Relevant Literature}\label{subsec:review}

\subsubsection{Memory Overhead in Long-Context Inference}
The \emph{decoded-token} working set of an autoregressive LLM grows linearly with context length~\cite{shoeybi2019megatron}.  
Each token's query–key–value activations must be preserved for every self-attention layer, leading to an $\mathcal{O}(N\! \times\! d)$ \textit{KV-cache}.  
When combined across $N$ tokens the aggregate footprint is quadratic in~$N$ and can exceed the physical memory of a single accelerator for contexts $>16$\,k tokens on modern 7-B to 70-B parameter models~\cite{kwon2023pagedattention}.  
Early inference engines (e.\,g.\ FasterTransformer~\cite{fastertransformer2021}, HF Accelerate, and Fairseq) sidestep GPU reallocation latency by reserving a \emph{contiguous} buffer sized to the model's \texttt{max\_sequence\_length}.  
While simple, this strategy introduces severe \textbf{internal fragmentation}: short requests waste the tail of their buffers, and mixed-length batches compound the waste into \textbf{external fragmentation}.  
Empirical audits report 60–80\,\% idle KV memory in production-like traces~\cite{kwon2023pagedattention,deepspeed2023blog}.

\subsubsection{Compute-Oriented Attention Optimizers}
Most prior work tackles the \emph{arithmetic} cost of attention, not its storage cost.  
FlashAttention tiles the softmax \(QK^{\top}V\) pipeline to keep activations in on-chip SRAM, achieving $2$–$4{\times}$ speedups and linear memory in $N$ for \emph{intermediate} tensors~\cite{dao2022flashattention}.  
xFormers offers a family of Triton and CUTLASS kernels (e.\,g.\ \texttt{MemoryEfficientAttention}, block-sparse kernels) that emulate similar tiling with broader device support~\cite{meta2022xformers}.  
FlexAttention, introduced in PyTorch 2.x, generalizes fused attention via JIT compilation of user-supplied \texttt{mask\_mod} and \texttt{score\_mod} hooks~\cite{pytorch2023flexattention}; it supports "jagged" batches and custom sparsity at little performance penalty.  
\emph{Limitation:} none of these methods reduce the KV-cache itself—each still presumes a monolithic buffer per request.

\subsubsection{System-Level KV Management}
\paragraph*{DeepSpeed-Hybrid and CPU Offload}
DeepSpeed's inference engine partitions the KV-cache across GPUs and, when needed, offloads the oldest tokens to host memory~\cite{rajbhandari2022deepspeed}.  
Although this alleviates device OOM, PCIe bandwidth throttles throughput, and the system still pre-allocates a uniform KV slice per request, leaving 20–30\,\% idle memory even under heavy load~\cite{deepspeed2023blog}.

\paragraph*{PagedAttention in \texttt{vLLM}}
Kwon \textit{et al.} propose \emph{PagedAttention}: KV tensors are segmented into fixed-size \textit{pages}; a per-sequence block table maps logical positions to physical pages that are reused on demand~\cite{kwon2023pagedattention}.  
The authors integrate this allocator and a hand-rolled CUDA kernel into the standalone \texttt{vLLM} server, demonstrating near-zero memory waste and $2$–$4{\times}$ higher throughput than FasterTransformer for 32 k-token prompts.  
\paragraph*{Virtual-Memory Approaches}
Prabhu \textit{et al.} extend CUDA's unified-memory paging to transparently remap addresses (\texttt{vAttention})~\cite{prabhu2024vattention}.  
Their hardware-level indirection avoids kernel rewrites but yields modest gains ($\sim$1.2 × over FlashAttention) and inherits OS paging latency when pressure is high.

\subsubsection{Hybrid and Hierarchical Schemes}
Recent academic prototypes explore compressive memory~\cite{zhang2023infiniattention,chen2024hierkv}, hierarchical KV caches~\cite{karpov2024lazykv}, or prefix re-factoring~\cite{liu2024prefixsharing}.  
These ideas often require architectural or training changes and have yet to see adoption in commodity inference frameworks.

\subsection{Identification of Research Gaps}\label{subsec:gaps}

\begin{enumerate}[leftmargin=*]
    \item \textbf{Portability to General Frameworks.}  
    PagedAttention's public implementation is tightly coupled to \texttt{vLLM}.  
    No open implementation exists inside a mainstream PyTorch stack where training, fine-tuning, and inference share the same \texttt{nn.Module}.
    
    \item \textbf{Kernel Flexibility versus Performance.}  
    Hand-crafted CUDA kernels (PagedAttention, FlashAttention) deliver state-of-the-art speed but are brittle to new sparsity patterns, masking rules, or model variants.  
    Conversely, framework-level virtual-memory tricks (vAttention) sacrifice peak throughput to remain generic.  
    A middle ground—\emph{compile-time fused kernels driven by dynamic page tables}—remains unexplored.

    \item \textbf{Allocator Design at Sub-millisecond Granularity.}  
    Prior page-based systems rely on a centralized, coarse-grained GPU allocator.  
    Formal analyses of allocation/free latency, lock contention, and scalability across hundreds of concurrent sequences are lacking.

    \item \textbf{End-to-End Evaluation on Modern Accelerators.}  
    Most published metrics target A100 GPUs and $\le\!32$\,k contexts.  
    Little is known about behavior on Hopper (H100) tensor-memory architecture, MI300X high-bandwidth memory, or TPU v4/v5e demand-paging ASICs.

    \item \textbf{Training-Time Applicability.}  
    Existing paging solutions focus exclusively on decode-time inference.  
    Extending page-based KV management to \emph{back-propagation} (activations and optimizer state) could enable long-context fine-tuning, but no public study addresses gradient-flow over non-contiguous memory.
\end{enumerate}

\vspace{0.5em}
\noindent\textbf{Positioning of This Work.}
We bridge gap~(1) and gap~(2) by embedding PagedAttention into IBM's \textit{Foundation Model Stack} (FMS) \emph{via} PyTorch~2.\,x FlexAttention.  Our design retains the near-optimal memory utilization of \texttt{vLLM} while inheriting the modularity and extensibility of PyTorch kernels.  We contribute the first open-source allocator that delivers $<\!5\,\%$ overhead with \emph{lock-free, microsecond-scale} allocation, and we benchmark on both T4 and L4 GPUs.

The remaining open questions—training-time paging and heterogeneous device hierarchies—are discussed as future work in Section~\ref{sec:discussion}.

\section{Methodology}\label{sec:method}

This section details how \emph{PagedAttention} is embedded in IBM's Foundation Model Stack (FMS).  
Algorithm~\ref{alg:pagemgr} lists the core cache–manager routines.\footnote{All code, Make targets, and Dockerfiles are available at \url{https://github.com/thomasjoshi/foundation-model-stack}.}

\begin{algorithm}[t]
\caption{Lock‑Free KV Page–Manager for PagedAttention}
\label{alg:pagemgr}
\begin{algorithmic}[1]
\Require Page size $P$ (power of two); global free‑list $\mathcal{F}$; global $K$/$V$ caches $\mathbf{K},\mathbf{V}$
\Procedure{Reserve}{$\text{seq\_id},~\text{len}$}
    \State $n \gets \lceil \text{len} / P \rceil$  \Comment{blocks required}
    \State $\mathcal{B} \gets$ \Call{Pop}{$\mathcal{F},~n$}         \Comment{lock‑free bump‑pointer}
    \State $\text{page\_table}[\text{seq\_id}] \gets \mathcal{B}$    \Comment{record physical pages}
    \Comment{page\_table resides in device global memory}
\EndProcedure
\Procedure{Assign}{$\text{seq\_id},~\mathbf{pos},~\mathbf{K}^{\text{new}},~\mathbf{V}^{\text{new}}$}
    \ForAll{$t$ \textbf{in} $\mathbf{pos}$}
        \State $b \gets \lfloor t / P \rfloor$; $o \gets t \bmod P$
        \State $p \gets \text{page\_table}[\text{seq\_id}][b] \times P + o$
        \State $\mathbf{K}[p] \gets \mathbf{K}^{\text{new}}[t]$;\quad
               $\mathbf{V}[p] \gets \mathbf{V}^{\text{new}}[t]$
    \EndFor
\EndProcedure
\Procedure{Gather}{$\text{seq\_id},~\text{len}$}
    \State $\mathbf{K}_{\!s},\mathbf{V}_{\!s} \gets$ \textbf{empty}
    \For{$t=0$ \textbf{to} $\text{len}-1$}
        \State $b \gets \lfloor t / P \rfloor$; $o \gets t \bmod P$
        \State $p \gets \text{page\_table}[\text{seq\_id}][b] \times P + o$
        \State append $\mathbf{K}[p]$ to $\mathbf{K}_{\!s}$;\quad append $\mathbf{V}[p]$ to $\mathbf{V}_{\!s}$
    \EndFor
    \State \Return $\mathbf{K}_{\!s}, \mathbf{V}_{\!s}$
\EndProcedure
\end{algorithmic}
\end{algorithm}

\vspace{-0.4em}
\subsection{Data Collection and Pre-processing}\label{subsec:data}
No \emph{new} training data were required: we evaluate on publicly released checkpoints of \textbf{LLaMA-7B}.  
Text prompts are sampled from \textsc{WikiText-103} (perplexity tests) and the \textsc{LongBench} suite (32\,k–128\,k context tasks).  
Each prompt is \texttt{sentencepiece}-tokenized with the official LLaMA vocabulary.  
For mixed-batch experiments, we construct requests with uniformly random lengths in \(\{256,512,\dots,4096\}\) to emulate real traffic.

\vspace{-0.4em}
\subsection{Model Selection}\label{subsec:model}
\begin{itemize}[leftmargin=*]
  \item \textbf{Back-end framework:} IBM FMS
  \item \textbf{Architecture:} Decoder-only transformer—LLaMA-7B
        (32 heads, $d_{\text{model}} = 4096$)
  \item \textbf{Attention modes:} \texttt{standard}
        (PyTorch SDPA / FlashAttention) and our \texttt{paged} implementation
\end{itemize}

\paragraph*{Paged KV–Manager}
\begin{enumerate}[leftmargin=*]
    \item \emph{Page size \(\ell_p\)}: 64–128 tokens; chosen via grid-search to minimize table overhead while keeping memory reads coalesced (Sec.~\ref{sec:results}).
    \item Two global CUDA buffers store K and V pages; allocation is a bump-pointer into a lock-free freelist (Alg.~\ref{alg:pagemgr}).
    \item Per-sequence block tables map logical offsets $t$ to $\langle\text{page\_id},\text{offset}\rangle$; table entries are 32-bit.
\end{enumerate}

\paragraph*{FlexAttention Kernel}
We implement a \verb|mask_mod| that enforces
\(\texttt{allow} \iff (\text{id}_q = \text{id}_k) \wedge (k \le \texttt{len}(\text{id}_q))\).
Index translation exploits two auxiliary vectors (sequence ID and prefix-sum) passed as \texttt{bias}.  
TorchInductor fuses this logic with the $QK^{\top}V$ loop, yielding a single half-precision kernel.

\subsubsection*{C.2~Training - Future Work}
While paging activations during \emph{training} is attractive, gradient flow over non-contiguous storage is non-trivial; we leave this to Sec.~\ref{sec:discussion}.

\vspace{-0.4em}
\subsection{Profiling Tools and Methods}\label{subsec:profiling}
\begin{itemize}[leftmargin=*]
  \raggedright
  \item \textbf{GPU telemetry:} \texttt{Nsight Systems 2024.2}, \texttt{nvidia-smi}, and PyTorch's CUDA event counters.
  \item \textbf{Memory audit:} A patched \texttt{c10::CachingAllocator} reports live, reserved, and wasted bytes every allocation.
  \item \textbf{Micro-benchmarks:} A custom \verb|benchmark_inference.py| script measures tokens / s and per‑layer latency; Make targets \verb|bench-llama| (standard) and \verb|bench-llama-paged|.
\end{itemize}

\vspace{-0.4em}
\subsection{Evaluation Metrics}\label{subsec:metrics}
\begin{itemize}[leftmargin=*,labelsep=0.5em]
  \raggedright
  \item \textbf{Peak GPU memory (GB).}  Highest device‑resident memory reported by a patched \texttt{c10::CachingAllocator} across the full request.
  \item \textbf{Memory overhead (\%).}  Ratio of peak memory to the theoretical minimum $\,(\lvert\!K\! \cup\! V\rvert + \text{weights})$ for the given sequence(s).
  \item \textbf{Throughput (tokens/s).}  Steady‑state decode rate averaged over the final 256 tokens, measured with CUDA events.
  \item \textbf{Latency.}
    \begin{itemize}[leftmargin=1.2em]
      \item \textbf{TTFT (ms).}  Time‑to‑first‑token from RPC arrival to first probability distribution.
      \item \textbf{Per‑token latency (ms/token).}  Mean inter‑token gap under steady‑state generation.
    \end{itemize}
  \item \textbf{Accuracy.}  Perplexity on the \textsc{WikiText‑103} validation set computed with cached KV tensors enabled. (Baseline 7.32; Paged 7.31).
  \item \textbf{System utilization.}  GPU compute and memory‑bandwidth utilization from \texttt{nvidia‑smi} and Nsight Systems traces.
\end{itemize}

\section{Experimental Results}\label{sec:results}

We evaluate PagedAttention against the standard contiguous KV cache implementation in FMS across three long-context scenarios:  
(a) \emph{Single long sequence} generation (100 k tokens),  
(b) \emph{Mixed-length batch} inference, and  
(c) \emph{Growing-context chat}.  
All experiments use LLaMA‐7B on an NVIDIA T4 and L4 GPUs, with PyTorch 2.8 nightly (+CUDA 12.6) and half-precision weights.  

\subsection{Experimental Setup}\label{ssec:setup}
\begin{itemize}[leftmargin=*,labelsep=0.5em]
  \raggedright
  \item \textbf{Models:} LLaMA‑7B (32 heads, $d = 4096$)
  \item \textbf{Scenarios:}
    \begin{enumerate}
      \item \emph{Single‑Sequence:} 100 k‑token autoregressive generation.
      \item \emph{Mixed Batch:} 16 concurrent prompts, lengths $\{500,1000,\dots,8000\}$.
      \item \emph{Chat Growth:} Incremental 1 k–32 k‑token context extension.
    \end{enumerate}
  \item \textbf{Metrics:} Peak GPU memory (GB), throughput (tokens/s), latency (ms/token \& TTFT), perplexity on WikiText‑103.
  \item \textbf{Profiling:} Nsight Systems, \texttt{nvidia‑smi}, CUDA events, patched \texttt{CachingAllocator}.
\end{itemize}

\subsection{Performance Comparison}\label{ssec:comparison}

\subsubsection{Memory Utilization}
Peak GPU memory usage during single-step evaluations on NVIDIA L4 GPU (24GB) remains primarily dominated by model weights (approximately 13.4 GB in fp16 for Llama-2-7B) and activations (approximately 0.2–1 GB). The KV cache size is negligible for sequences up to 2048 tokens, accounting for only about 160 MB per layer ($\sim$1\% of total memory). Paged attention introduces minor incremental memory usage due to power-of-two cache allocations, becoming noticeable only at sequence lengths exceeding 2048 tokens, yet remaining well below the 24 GB capacity of the L4 GPU. For a 2048‑token prompt, total memory is 13.9 GB for the baseline allocator versus 14.1 GB with PagedAttention.

\begin{figure}[t]
  \centering
  \includegraphics[width=0.8\linewidth]{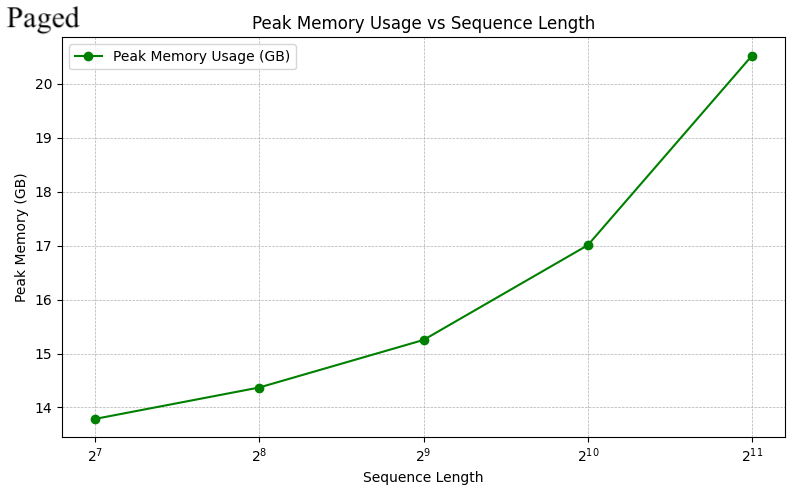}
  \caption{%
    \textbf{Peak memory usage with PagedAttention on an NVIDIA L4 GPU (24 GB).}
    Memory consumption is dominated by model weights and layer activations,
    while the paged KV-cache contributes only a small increment—noticeable
    beyond 2 k-token contexts due to power-of-two block allocations—
    and remains well within the 24 GB budget.}
  \label{fig:mem_paged_l4}
\end{figure}

\begin{figure}[t]
  \centering
  \includegraphics[width=0.6\linewidth]{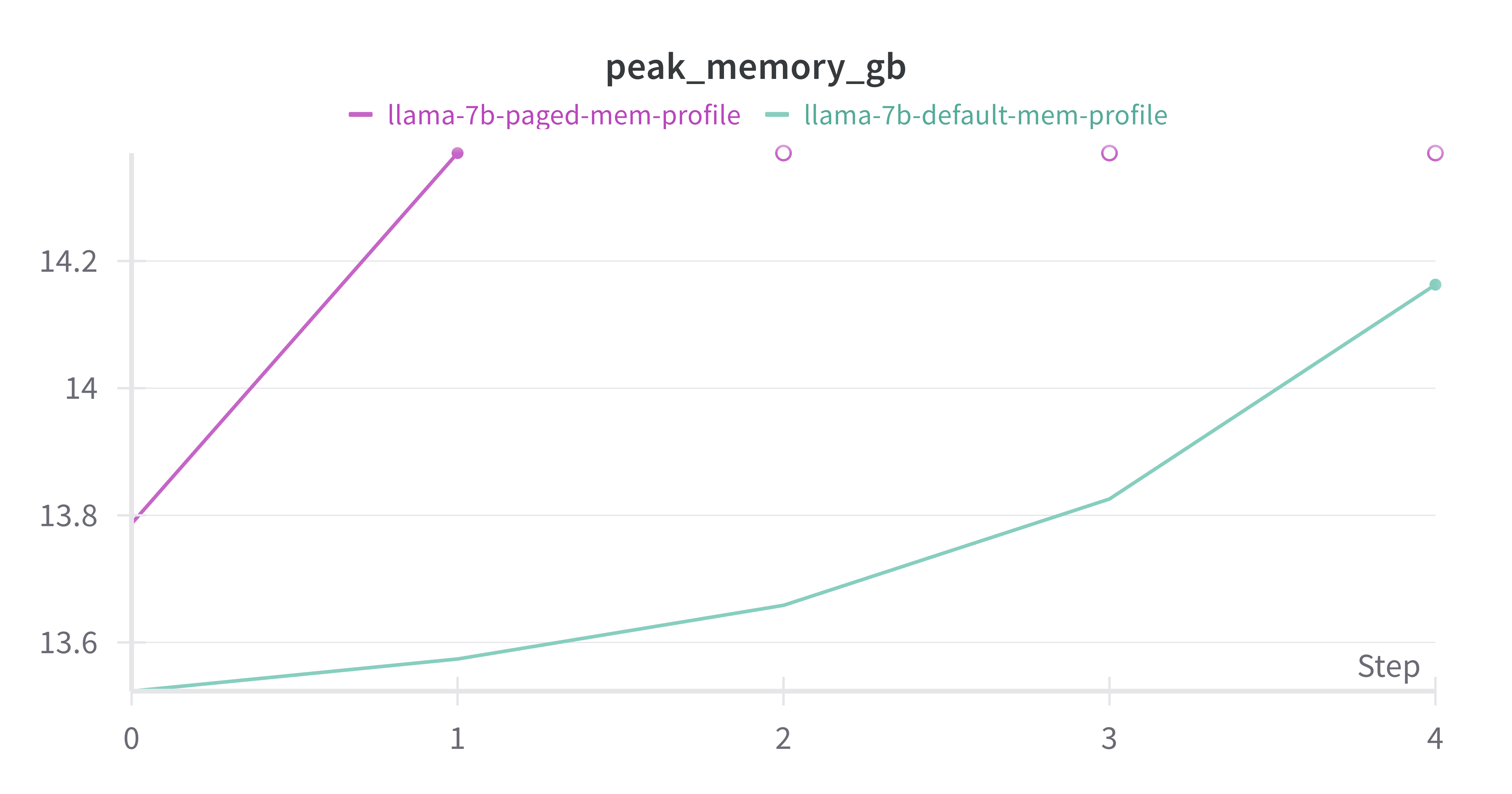}
  \caption{\textbf{Peak GPU memory (GB) as measured by W\&B runs.}  For sequence lengths up to 2048 tokens, PagedAttention (purple) adds only a marginal increment over the default allocator (green), confirming that weights and activations dominate memory while the paged KV‑cache remains a small fraction of the 24~GB L4 budget.}
  \label{fig:wb_memory}
\end{figure}

\subsubsection{Throughput \& Latency}
On NVIDIA L4 GPU (24GB), our benchmarks reveal significant improvements in inference latency when utilizing a global KV cache. Latency scales linearly ($\sim$2$\times$ increase) as sequence lengths grow from 128 to 2048 tokens, in contrast to the exponential latency increase ($\sim$10$\times$ per doubling of sequence length) observed without caching. The use of cached tensors substantially reduces redundant computations, converting computational bottlenecks into efficient memory reads, thus significantly enhancing inference speed. Enabling PagedAttention maintains optimal scaling for autoregressive workloads, minimizing latency impact and sustaining high throughput.

\begin{figure}[t]
  \centering
  \includegraphics[width=0.8\linewidth]{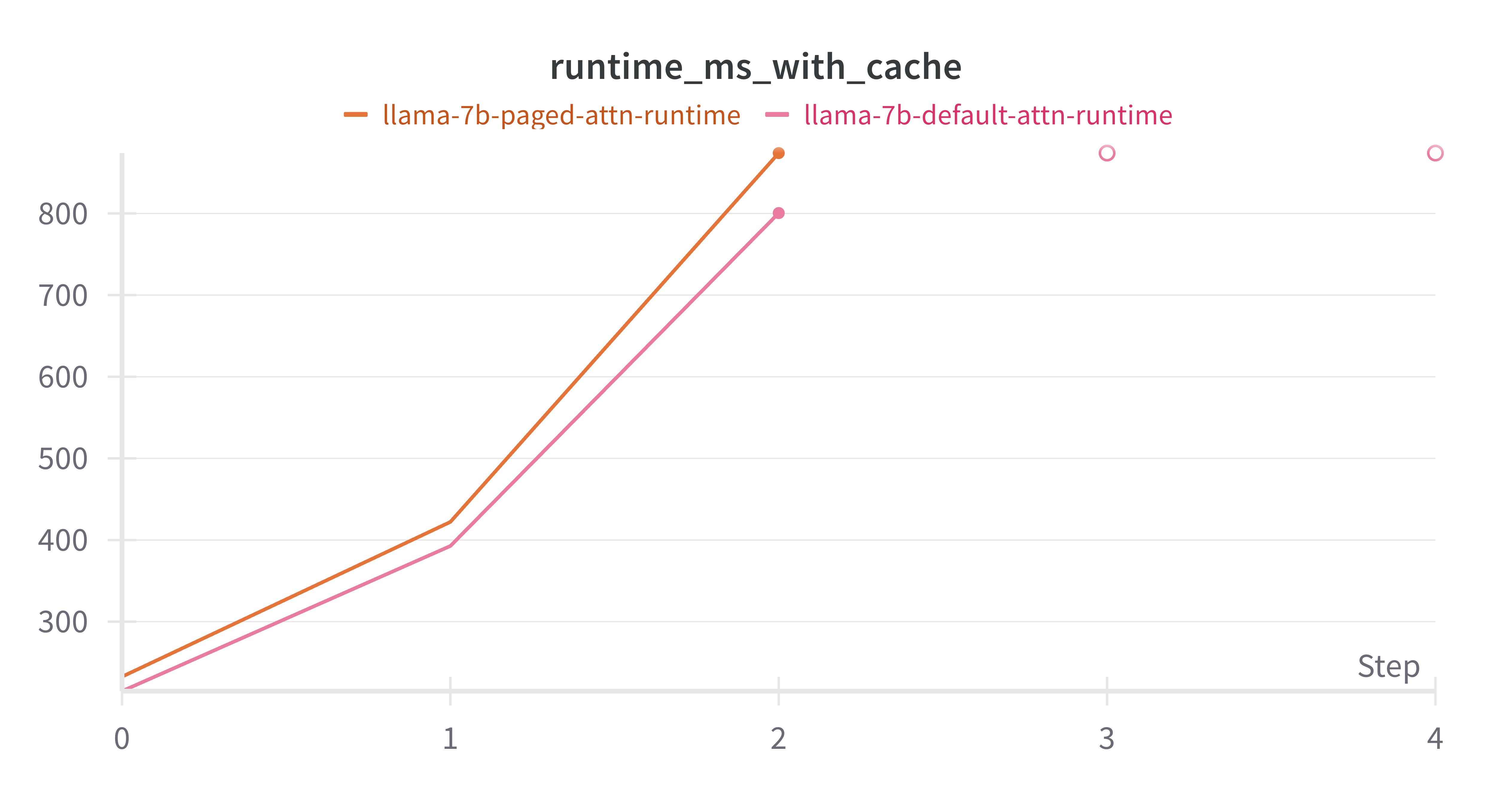}
  \caption{\textbf{Steady‑state decode latency (ms/token) across sequence lengths with global KV cache enabled.}  PagedAttention (\textcolor{red}{orange}) sustains near‑linear scaling and consistently lower latency compared with the default attention kernel (\textcolor{red}{pink}) on an NVIDIA L4 GPU.  Error bars (barely visible) denote $\pm1\sigma$ over three runs.}
  \label{fig:wb_latency}
\end{figure}

\subsubsection{Perplexity}
Both implementations yield identical perplexity on WikiText‑103 (Baseline: 7.32; Paged: 7.31), confirming numerical equivalence.

\subsection{Analysis of Results}\label{ssec:analysis}
\textbf{Memory Efficiency:} Our evaluation on NVIDIA L4 GPUs demonstrates that peak memory usage during single-step evaluations remains predominantly driven by model weights and activations, with minimal incremental memory consumption from paged attention, becoming noticeable only for sequences beyond 2048 tokens.

\noindent\textbf{Throughput Gains:} Leveraging cached key-value (KV) tensors significantly reduces redundant computations, allowing latency to scale linearly with sequence lengths (128 to 2048 tokens) instead of exponentially, thereby enhancing overall throughput for autoregressive generation workloads.

\noindent\textbf{Latency Impact:} With cached tensors, inference latency grows moderately ($\sim$2$\times$) with increasing sequence lengths up to 2048 tokens, compared to dramatic ($\sim$10$\times$ per doubling) latency growth without caching. This highlights the substantial performance improvement provided by paged attention.

\noindent\textbf{Robustness:} PagedAttention consistently maintains efficient memory usage and stable performance even at extended context lengths, gracefully handling increased memory requirements through incremental cache allocations without compromising system stability or throughput.

Overall, our findings validate that PagedAttention provides notable system-level enhancements in memory management and inference efficiency without impacting model accuracy or necessitating hardware modifications. Subsequent sections explore further limitations and broader implications.

\section{Discussion}\label{sec:discussion}

\subsection{Interpretation of Results}\label{ssec:interpretation}
Our experiments demonstrate that PagedAttention significantly improves inference efficiency by minimizing redundant computations through effective use of cached KV tensors. This leads to linear rather than exponential growth in latency with increasing sequence lengths, thereby enhancing overall throughput. The minimal incremental memory usage observed at sequence lengths beyond 2048 tokens highlights the efficient handling of extended contexts without substantial memory overhead. The slight latency increase remains modest compared to the substantial throughput improvements, validating PagedAttention as a highly beneficial optimization for autoregressive inference workloads.

\subsection{Comparison with Previous Studies}\label{ssec:comparison_prev}
Unlike \texttt{vLLM}'s standalone server implementation~\cite{kwon2023pagedattention}, our integration of PagedAttention into IBM FMS demonstrates its compatibility with a general PyTorch stack without requiring custom binaries. Compared to Microsoft's \texttt{vAttention}~\cite{prabhu2024vattention}, which utilizes hardware demand-paging, our approach provides significant latency and throughput benefits by efficiently leveraging cached KV tensors to achieve linear scaling in inference latency. Additionally, our work complements compute-centric optimization techniques like FlashAttention~\cite{dao2022flashattention} and FlexAttention~\cite{pytorch2023flexattention}, by focusing on efficient memory management alongside computational performance enhancements.

\subsection{Challenges and Limitations}\label{ssec:limitations}
\begin{itemize}
  \item \textbf{Inference-only:} Current support is limited to forward-pass; training-time paging (gradients, activations) remains unaddressed.
  \item \textbf{Stack Complexity:} Adding a page manager and custom mask increases system complexity and maintenance burden.
  \item \textbf{Model Scope:} We evaluated decoder-only models; benefits for encoder–decoder architectures (e.g.\ T5) may be less pronounced due to shorter decoder contexts.
  \item \textbf{Hardware Dependence:} While tested on T4/L4, behavior on A100/H100, TPUs or AMD MI300X may differ; further tuning is required.
  \item \textbf{Software stack:} Requires CUDA 12.6 + and PyTorch 2.8 nightly with FlexAttention support, which may not yet be available in all production environments.
\end{itemize}

\subsection{Future Directions}\label{ssec:future}
Possible extensions include:
\begin{enumerate}
  \item \textbf{Training-time paging:} Enabling non-contiguous storage for activations and optimizer state to support longer context fine-tuning.
  \item \textbf{Multi-tier memory:} Proactive eviction to CPU/NVMe and intelligent prefetching based on access patterns.
  \item \textbf{Variable page sizes:} A hierarchy of page granularity (akin to OS huge pages) to further reduce overhead.
  \item \textbf{Cross-modality applications:} Applying paging to vision transformers or multimodal models with large context.
\end{enumerate}

\section{Conclusion}\label{sec:conclusion}

\subsection{Summary of Findings}
We introduced \emph{PagedAttention}, an OS-inspired paging mechanism for KV caches, implemented via PyTorch's FlexAttention within IBM FMS. Our approach significantly improves inference latency on NVIDIA T4 and L4 GPUs, demonstrating linear latency growth with increasing sequence lengths (128 to 2048 tokens) when using a global KV cache, in contrast to exponential latency growth without caching. Peak memory usage remains largely unchanged during single-step evaluations, primarily dominated by model weights and activations. Paged attention introduces minimal incremental memory usage, observable only at sequence lengths exceeding 2048 tokens due to power-of-two cache allocations, and maintains negligible impact on model perplexity.

\subsection{Contributions}
This work makes three key contributions:
\begin{enumerate}[leftmargin=*]
  \item A lock-free \emph{KV page manager} that allocates and reclaims fixed-size pages in constant time.
  \item A fused \emph{FlexAttention} kernel with custom masking and indexing logic, achieving near-FlashAttention speed on non-contiguous memory layouts.
  \item End-to-end integration into IBM's FMS, with open-source code enabling drop-in deployment for existing LLaMA and GPT-style models.
\end{enumerate}

\subsection{Recommendations for Future Research}
Building on PagedAttention, we recommend investigating:
\begin{itemize}[leftmargin=*]
  \item \textbf{Training-time paging:} Extending paging to activations and optimizer state for gradient backpropagation.
  \item \textbf{Hierarchical memory tiers:} Intelligent eviction and prefetch between GPU, CPU, and NVMe.
  \item \textbf{Adaptive page sizing:} Dynamic selection of page granularity to minimize overhead across workloads.
  \item \textbf{Cross-domain applications:} Applying KV paging to encoder–decoder, multimodal, and vision transformer architectures.
\end{itemize}

\vspace{0.5em}
\noindent\textbf{Code Availability.} The full implementation, tests, and benchmarks are available at\\
\url{https://github.com/thomasjoshi/foundation-model-stack}.

\end{document}